\documentclass{article}
\usepackage{arxiv}
\usepackage[utf8]{inputenc} % allow utf-8 input
\usepackage[T1]{fontenc}    % use 8-bit T1 fonts
\usepackage{hyperref}       % hyperlinks
\usepackage{url}            % simple URL typesetting
\usepackage{booktabs}       % professional-quality tables
\usepackage{amsfonts}       % blackboard math symbols
\usepackage{nicefrac}       % compact symbols for 1/2, etc.
\usepackage{microtype}      % microtypography
\usepackage{lipsum}		% Can be removed after putting your text content
\usepackage{graphicx}

\usepackage{xcolor}
\usepackage{dirtytalk}
\usepackage{soul} %Use \st
\usepackage{amsmath}

\title{Differential Replication in Machine Learning}

\author{\href{https://orcid.org/0000-0002-7422-1493}{\includegraphics[scale=0.06]{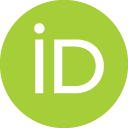}\hspace{1mm}Irene ~Unceta}\\
	BBVA Data \& Analytics\\
	Department of Mathematics and Computer Science\\
	Universitat de Barcelona\\
	\texttt{irene.unceta@bbvadata.com} \\
	\And
	\href{https://orcid.org/0000-0002-9659-2762}{\includegraphics[scale=0.06]{orcid.png}\hspace{1mm}Jordi ~Nin} \\
	Department of Operations, Innovation and Data Sciences\\
	Universitat Ramon Llull, ESADE\\
	\texttt{jordi.nin@esade.edu} \\
    \AND
	\href{https://orcid.org/0000-0001-7573-009X}{\includegraphics[scale=0.06]{orcid.png}\hspace{1mm}Oriol ~Pujol} \\
	Department of Mathematics and Computer Science\\
	Universitat de Barcelona\\
	\texttt{oriol\_pujol@ub.edu}\\
}

\date{}

\hypersetup{
    pdftitle={Differential Replication in Machine Learning},
    pdfauthor={Irene ~Unceta, Jordi ~Nin, Oriol ~Pujol},
    pdfkeywords={Differential Replication, Machine Learning Copies, Self-evolving Systems},
}

\begin{document}

\maketitle

\begin{abstract}
When deployed in the wild, machine learning models are usually confronted with data and requirements that constantly vary, either because of changes in the generating distribution or because external constraints change the environment where the model operates. To survive in such an ecosystem, machine learning models need to adapt to new conditions by evolving over time. The idea of model adaptability has been studied from different perspectives. In this paper, we propose a solution based on reusing the knowledge acquired by the already deployed machine learning models and leveraging it to train future generations. This is the idea behind differential replication of machine learning models.
\end{abstract}

\section{Survival of the fittest}\label{sec:introduction}

\say{\textit{If during the long course of ages and under varying conditions of life, organic beings vary at all in the several parts of their organization,  [...] I think it would be a most extraordinary fact if no variation ever had occurred useful to each being's own welfare, in the same way as so many variations have occurred useful to man. But if variations useful to any organic being do occur, assuredly individuals thus characterized will have the best chance of being preserved in the struggle for life; and from the strong principle of inheritance they will tend to produce offspring similarly characterized. This principle of preservation, I have called, for the sake of brevity, Natural Selection.}} [Charles Darwin, Origin of the Species, p.127, 1859]\\

Natural Selection explores how organisms adapt to a changing environment in their struggle for survival\cite{darwin}. Survival in this context is intrinsically defined by a complex and generally unknown fitness function that governs the life of all living creatures. The closer they move towards the optimal value of this function, the better fit they are to face the hard conditions imposed by their environment and, hence, the better chance they have at survival. 

This pressure imposed by the environment is not unique to living organisms. It is also present in artifacts of human culture, business and society, including everything from economic changes, adjustment of moral and ethical concerns, regulatory revisions or the reframing of societal rules that results from unexpected global crises. In a smaller scale, it is also present in machine learning model deployment. 

Machine learning models are only a part of the larger ecosystem entailed by a machine learning system. A machine learning system comprises all the elements that interact with a model throughout its lifespan. These include the data and their different sources, the deployment infrastructure, the governance protocol, or the general regulatory framework. Components of a machine learning system may be both internal and external to the company. This means that they are often out of the data practitioner's control. Plus, given the variable nature of most of its external components, a machine learning system's environment is prone to change in time. The Gartner Data Science Team Survey ~\cite{Gartner2019} found that over 60\% of machine learning models developed in companies are never actually put into production, due mostly to a lack of alignment with environmental demands. To survive in such an ecosystem, machine learning models need to adapt to new conditions by learning to evolve over time.

This notion of adaptability has been present in the literature since the early times of machine learning, as practitioners have had to devise ways in which to adapt theoretical proposals to their everyday life scenarios \cite{barocas2017, veale2017, kroll2018}. As the discipline has evolved, so have the available techniques to this end. Consider, for example, situations where the underlying data distribution changes resulting in a concept drift. Traditional batch learners were incapable of adapting to such drifts. Online learning algorithms were devised~\cite{bottou2004} to succeed in this task by iteratively updating their knowledge according to the data shifts. In other situations, it is not the data that change but the business needs themselves. For instance, fraud detection algorithms~\cite{deepfraud2017} are regularly retrained to incorporate new types of fraud. Commercial machine learning applications are designed to answer very specific business objectives that may evolve in time. Take, for example, the case where a company wants to focus on a new client portfolio. This may require evolving from a binary classification setting to a multi-class configuration~\cite{escalera2009}. Under such circumstances, the operational point of a trained classifier can be changed to adapt it to the new policy. Alternatively, it is also possible to add patches in the form of wrappers to already deployed models. These endow models with new traits or functionalities that help them adapt to the new data conditions, either globally\cite{mena2019} or locally\cite{ribeiro2016}.

In contrast, when it is the very nature of the system that changes, alternatives are scarce. When change affects the entire ecosystem of a model, most existing solutions are not applicable. Say, for example, that one of the original input attributes is no longer available, that a devised black-box solution is required to be interpretable or that updated software licenses require moving to a new production environment. It is such drastic changes in the demands of a machine learning system that this article is concerned with.

A straightforward solution in this context is to discard the existing model and re-train a new one. However, by discarding the given solution, we loose all the acquired knowledge and have to rebuild and validate the full machine learning stack. This is seldom the most efficient not the most effective way for tackling this challenge. In this paper, we explore this problem and advocate for imitating the way in which biological systems adapt to changes. In particular, we stress the importance of reusing the knowledge acquired by the already existing machine learning model in order to train a second generation that is better adapted to th enew conditions. 

\section{Modelling adaptation to new environments}

Adaptation of machine learning models to new scenarios or domains is a well known problem in the machine learning literature. The most well known research branches are transfer learning and domain adaptation. Domain adaptation usually deals with changes in the data distributions as time evolves, or with learning knowledge representations for one domain such that they can be transferred to another related target domain. For example, due to the COVID-19 pandemic, several countries decided to accept card payments without introducing the pin up to 50 euros instead of the previous 20 euros limit, in order to minimize the interactions of card holders with the points of selling. This domain modification may affect to the fraud card detection algorithms, requiring modifications in their usage. 

Conversely, transfer learning refers to the specific case where the knowledge acquired when solving one task is recycled to solve a different, yet related task~\cite{pan2010}. In general, the problem of transfer learning can be mathematically framed as follows. Given source and target domains $\mathcal{D}_s$ and $\mathcal{D}_t$ and their corresponding tasks $\mathcal{T}_s$ and $\mathcal{T}_t$, the goal of transfer learning is to build a target conditional distribution $P(y_t|x_t)$ in $\mathcal{D}_t$ for task $\mathcal{T}_t$ from the information obtained when learning $\mathcal{D}_s$ and $\mathcal{T}_s$, where $\mathcal{T}_s \neq \mathcal{T}_t$ and $\mathcal{D}_s \neq \mathcal{D}_t$. Advantages of this kind of learning when compared with traditional learning are that learning is performed much faster, requiring less data, and even achieving better accuracy results.

In this work we focus in a different adaptation problem. In our described scenario, the task remains the same but the existing solution is no longer fit because of changes in the environmental constraints. We can frame the problem at hand using the former notation as follows. Given a domain $\mathcal{D}_s$, its corresponding task $\mathcal{T}$, and the set of original environmental constraints $\mathcal{C}_s$ that make the solution of this problem feasible we assume an scenario were a hypothesis space $\mathcal{H}_s$ has been defined. In this context, we want to learn a new solution for the same task and for a new target scenario defined by the set of feasibility constraints $\mathcal{C}_t$, where $\mathcal{C}_t \neq \mathcal{C}_s$. This may or may not require the definition of a new hypothesis space $\mathcal{H}_t$. In a concise form and considering an optimization framework this can be rewritten as
\begin{center}
\begin{tabular}{ccc}
\begin{tabular}{ll}
\textit{Scenario I}\\
$\text{for} \ \mathcal{T} \ \text{in}\; \mathcal{D}_s$\\
\\
$\underset{\tiny \text{for} \; h\in \mathcal{H}_s}{\text{maximize}}$ & $P(y|x;h)$\\
subject to &$\mathcal{C}_s$ \\

\end{tabular}
& 
$\rightarrow$
&
\begin{tabular}{ll}
\textit{Scenario II}\\
$\text{for} \ \mathcal{T} \ \text{in}\; \mathcal{D}_t$\\
\\
$\underset{\tiny for \; h\in \mathcal{H}_t}{\text{maximize}}$ & $P(y|x;h)$\\
subject to &$\mathcal{C}_t$\\
\end{tabular}
\end{tabular}
\end{center}

This problem corresponds to that of \emph{environmental adaptation}. Under this notation, the original solution corresponds to a model $h_s$ that belongs to the hypothesis space $\mathcal{H}_s$ defined for the first scenario. This is a model that fulfills the constraints $\mathcal{C}_s$ and maximizes $P(y|x;h)$ for a training dataset $\mathcal{S}=\{(x,y)\}$, defined by task the $\mathcal{T}$ on the domain $\mathcal{D}_s$. Adaptation involves transitioning from this original scenario to \textit{Scenario II}; a process which may be straightforward. Although, in general, this is not the case. 

The solution obtained for the first scenario may be unfeasible in the second; thus, effectively ceasing the lifespan of the learned model. This happens when $h_s$ is outside of the feasible set defined by $\mathcal{C}_t$. In the most benevolent case, there may exist another solution from the original hypothesis space, $\mathcal{H}_s$, that fulfills the new constraints. However, there might be no overlapping between the set of constraints $\mathcal{C}_t$ and the set of models defined by $\mathcal{H}_s$. This implies that no model of that hypothesis space can be considered a solution. In such cases, adaptation involves definition of a new hypothesis space $\mathcal{H}_t$ altogether. Take, for example, the case of an application with a multivariate Gaussian kernel support vector machine. Assume that due to changes in the existing regulation, models are required to be fully interpretable in the considered application. The new set of constraints is not compatible with the original scenario and hence we would require a complete change of substrate. 

In this scenario, we introduce the notion of differential replication of machine learning models as an efficient approach to ensure model survival in a new demanding environment, by building on knowledge acquired in previous generations. This is solving for \textit{scenario II} considering the solution obtained for \textit{Scenario I}. Differential replications ensures the environmental adaptation of machine learning models when they are subjected to constant changes.

\section{Differential replication}

\say{\textit{When copies are made with variation, and some variations are in some tiny way "better" (just better enough so that more copies of them get made in the next batch), this will lead inexorably to the ratcheting process of design improvement Darwin called evolution by natural selection.}} [Daniel Dennett, Breaking the Spell, p.78, 2006] \\

\noindent
Solving the former \emph{environmental adaptation} problem can sometimes be straightforwardly done by discarding the existing model and re-training a new one. This is possible when the new model hypothesis space $\mathcal{H}_t$ is required to provide feasible solutions for the constraint set $\mathcal{C}_t$. However, it is worth considering the costs of this approach. In general, rebuilding a model from scratch (i) implies obtaining the clearance from legal, business, ethical, and engineering departments, (ii) does not guarantee that a good or better solution of the objective function will be achieved\footnote{The objective function in this scenario corresponds to $P(y|x;h)$.}, (iii) requires a whole new iteration of the machine learning pipeline, which is costly and time-consuming, (iv) assumes full access to the training dataset, which may no longer be available or require a very complex version control process. Plus, in many companies machine learning solutions are kept up-to-date using automated systems that continuously evaluate and retrain models; a technique known as continuous learning. Note, however, that this may take huge storage space, due to the need to save all the new incoming information. Hence, in the best case scenario, re-training is an expensive and difficult approach that assumes a certain level of knowledge that is not always guaranteed. Nonetheless, this is the most commonly used technique for solving this environmental adaptation problem. In what follows we consider other techniques.

Environmental adaptation is well known for living beings. Under the theory of Natural Selection, adaptation relies on changes in the phenotype of a species over several generations to guarantee its survival and evolution. This is sometimes referred to as \emph{differential reproduction}. In the same lines, we define \emph{differential replication} of a machine learning model as a cloning process in which traits are inherited from generation to generation while at the same time adding variations that make descendants more suitable/fit to the new environment.  

More formally, differential replication refers to the process of finding a solution $h_t$ that fulfills the constraints $\mathcal{C}_t$, i.e. it is a feasible solution, while preserving/inheriting features from $h_s$. Note that, in general, $P(y|x;h_t)\sim  P(y|x;h_s)$. In the best case scenario, we would like to preserve or improve the performance of the source solution $h_s$, also known as the parent. However, this is requirement that may not always be achieved. In a biological simile, requiring a guepard to be able to fly may imply loosing its ability to run fast. 

\section{Differential replication mechanisms}

In this section, we consider existing approaches to implement \emph{differential replication} in its attempt to solve the problem of \emph{environmental adaptation}. 

%\begin{figure}
%    \centering
%    \includegraphics[width=\linewidth]{spectrum.png}
%    \caption{Inheritance spectrum and techniques.}
%    \label{fig:my_label}
%\end{figure}

\begin{figure}
    \centering
    \includegraphics[width=0.7\linewidth]{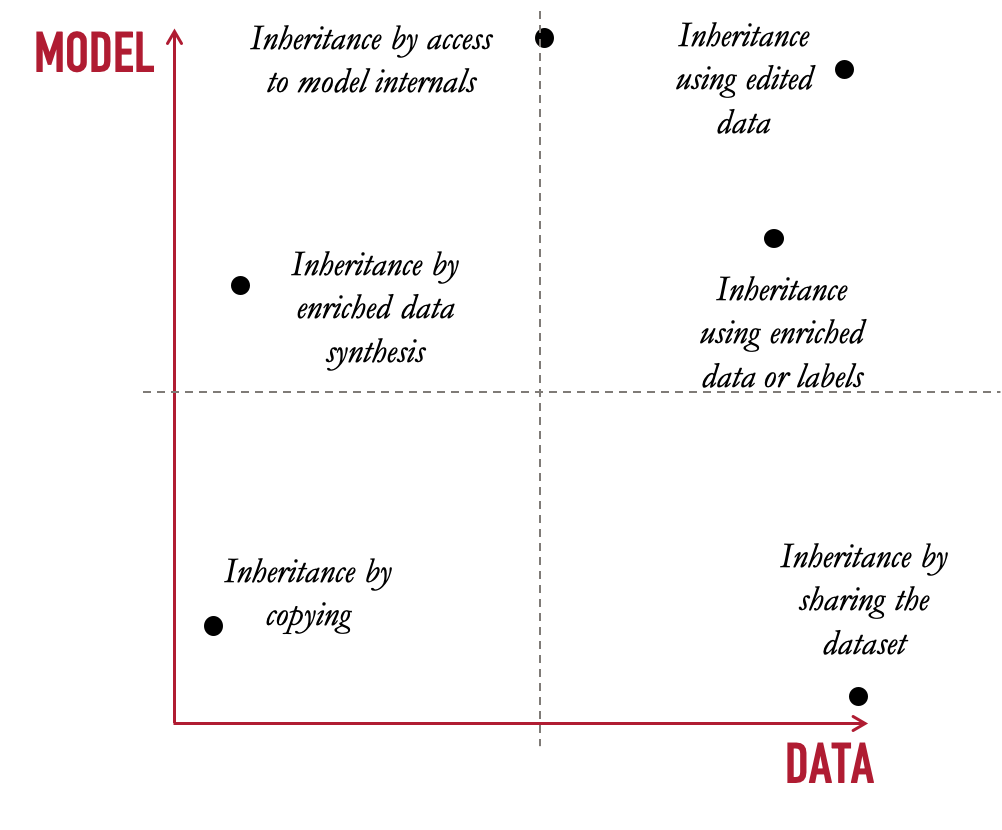}
    \caption{Graphical representation of inheritance mechanisms in terms of their knowledge of the data and the model internals.}
    \label{fig:my_label}
\end{figure}

The notion of \emph{differential replication} is built on top of two concepts. First, there is some inheritance mechanism that is able to transfer key aspects from the previous generation to the next. That would account for the name of \emph{replication}. Second, the next generation should display new features or traits not present in their parents. This corresponds to the idea of \emph{differential}. These new traits should make the new generation more fit to current environment to enable \emph{environmental adaptation} of the offspring.

Particularizing to machine learning models, implementing the concept of \emph{differential} may involve a fundamental change in the substratum of the given model. This means we need a new hypothesis space that will ensure that part of the models in that space fulfill the constraints of the new environment $\mathcal{C}_t$. Consider, for example, the case of a large ensemble of classifiers. In highly time demanding tasks, this model may be too slow to provide real time prediction when deployed into production. Differential replication of this model enables moving from this architecture to a simpler, more efficient one, such as that of a shallow neural network~\cite{bucilua2006}. This "child" network can inherit the decision behavior of its predecessor while at the same time being more fit to the new environment.

Conversely, \emph{replication} requires that some behavioral aspect be inherited by the next generation. Usually, it is the model's decision behavior that is inherited, so that the next generation will replicate the parent decision boundary. Replication can be attained in many different ways. As shown in Fig. \ref{fig:my_label}, depending on the amount of knowledge that is assumed about the initial data and model, mechanisms for inheritance can be categorized as follows:

\begin{itemize}
    \item {\bf Inheritance by sharing the dataset:} Two models trained on the same data are bound to learn similar decision boundaries. This is the weakest form of inheritance possible, were no actual information is transferred from source to target. Here the decision boundary is reproduced indirectly and mediated through the data themselves. Re-training falls under this category. This form of inheritance requires no access to the parent model, but assumes knowledge of its training data.

    \item {\bf Inheritance using edited data:} Editing is the methodology that allows data selection for training purposes\cite{bhattacharya1981,bhattacharya2005,mukherjee2004}. Editing can be used to preserve those data that are relevant to the decision boundary learned by the original solution and use them to train the next generation. Take, for example, the case where the source hypothesis space corresponds to the family of support vector machines. In training a differential replica, one could retain only those data points that were identified as support vectors. This mechanism assumes full access to the model internals, as well as to the training data.
    
    \item {\bf Inheritance using model driven enriched data:} Data enrichment is a form of adding new information to the training dataset through either the features or the labels. In this scenario, each data sample in the original training set is augmented using information from the parent decision behavior. For example, a sample can be enriched by adding additional features using the prediction results of a set of classifiers. Alternatively, if instead of learning hard targets one considers using the output of the parent's class probability outputs or logits as soft-targets, this richer information can be exploited to build a new generation that is closer in behavior to the parent. Under this category fall methods like model distillation \cite{bucilua2006, hinton2015, szegedy2016, yang2018}, as well as techniques such as label regularization\cite{muller2019, yuan2020} and label refinery\cite{bagherinezhad2018}. In general, this form of inheritance requires access to the source model and is performed under the assumption of full knowledge of the training data. 
    
    \item{\bf Inheritance by enriched data synthesis} A similar scenario is that where the original training data is not accessible, but the model internals are open for inspection. In this situation, the use of synthetic datasets has been explored\cite{bucilua2006, zeng2000}. In some cases, intermediate information about the representations learned by the source model are also used as a training set for the next generation. This form of inheritance can be understood as a zero-shot distillation\cite{nayak2019}.
    
    \item {\bf Inheritance of internals model's knowledge: } In some cases, it is possible to access the internal representations of the parent model, so that more explicit knowledge can be used to build the next generation \cite{aguilar2019, cheng2020}. For example, if both parent and child are neural networks, one can force the mid-layer representations to be shared among them\cite{uncetacomplex}. Alternatively, one could use the second level rules of a decision tree to guide the next generation of rule-based decision models.
    
    \item{\bf Inheritance by copying} In highly regulated environments, access to the original training samples or to the model internals is not possible. In this context, experience can also be transmitted from one model to its differential replica using synthetic data points labelled according to the hard predictions of the source model. This has been referred to as copying\cite{uncetaarxiv}.
\end{itemize}

Note that on top of a certain level of knowledge about either the data or the model, or both, some of the techniques listed above often impose also additional restrictions on the considered scenarios. Techniques such as distillation, for example, assume that the original model can be controlled by the data practitioner, i.e. internals of the model can be tuned to force specific representations of the given input throughout the adaptation process. In certain environments this may be possible, but generally it is not.

\section{Differential replication in practice}

To illustrate the utility of differential replication, we describe six different scenarios where it can be exploited to ensure a devised machine learning solution adapts to different changes in its environment.

\subsection*{Intelligible explanations of non-linear phenomena}

A widely established technique to provide explanations is to use linear models, such as logistic regression. Model parameters, i.e. the linear coefficients associated to the different attributes, can be exploited to provide explanations to different audiences. Although this approach works in simple scenarios where the variables do not need to be modified nor pre-processed, this is never the case for real life applications, where variables are usually redesigned before training and new more complex features are often introduced. This is even worse when, in order to improve model performance, data scientists create a large set of new variables, such as bi-variate ratios or logarithm scaled variables, to capture non-linear relations between original attributes that linear models cannot handle during the training phase. This results in new variables being obfuscated and therefore often not intelligible for humans.

A straightforward solution using differential replication is to replace the whole predictive system, composed by both the pre-processing/feature engineering step and the machine learning model by a copy that considers both steps as a single black box model\cite{uncetanips}. Doing this, we are able to deobfuscate model variables by training copies to learn the decision outputs of trained models directly from the raw data attributes without any pre-process.

\subsection*{Moving to a different software environment}

In-company infrastructure is subject to continuous updates due to the rapid pace with which new software versions are released to the market. Changes in the organizational structure of a company may drive the engineering department to change course. Say, for example, that a company whose products were originally based on Google's Tensorflow package \cite{tensorflow} makes the strategic decision of moving to Pytorch\cite{pytorch}. In doing so, they might decide to re-train all models from scratch in the new environment. This is a long and costly process that can even result in a loss of performance. Specially if original data is not available or the in-house data scientist are new to this framework. Alternatively, using differential replication, the knowledge acquired by the existing solutions could be exploited in the form of hard or soft labels or as additional data attributes for the new generation. 

\subsection*{Mitigating bias in prediction}

As a model is tested against new data throughout its lifespan, some of its learned biases may be made apparent. Under such circumstances, one may wish to transit to a new model that inherits the original predictive performance but which ensures non-discriminatory outputs. A possible option is to edit the sensitive attributes to remove any bias, reducing in this way the disparate impact in the task $\mathcal{T}$, and then training a new model on the edited dataset. Alternatively, in very specific scenarios where the sensitive information is not leaked through additional features, it is possible to build a copy by removing the protected data variables\cite{uncetaibpria}. Or even, redesign the hypothesis space considering a loss function to account for the fairness dimension when training subsequent generations.

\subsection*{Concept drift in batch learning}

Batch machine learning models are rendered obsolete by their inability to adapt to a change in the data distribution. When this happens the most straightforward solution is to wait until there are enough samples of the new distribution and re-train the model. However, this solution is timely and often expensive. A faster solution is to use the idea of differential replication to create a new enriched dataset able to detect the data drift. For example, including the soft targets and a timestamp attribute in the target domain, $\mathcal{D}_t$. Then, train a new model using this enriched dataset that replicates the decision behavior of the previously trained classifier. Finally, we also need to allow the new model to accept new incoming data samples to be learned. This second characteristic can be added by incorporating the online requirement in the $\mathcal{C}_t$ constrains for the differential replication process\cite{uncetaccia}. 

\subsection*{Privacy preserving models}

Developing good machine learning models requires abundant data. The more accessible the data, the more effective a model will be. In real applications, training machine learning models usually requires collecting a large volume of data from users, often including sensitive information. Directly releasing these models trained using user data could derive in privacy breaches, as the risk of leaking sensitive information encoded in the public model increases.

Differential replication can be used to avoid this issue by is training another model, usually a simpler one, that replicates the learned decision behavior but which preserves the privacy of the original training set by not being directly linked to these data. The use of distillation techniques in the context of teacher-student networks, for example, has been reported to be successful in this task~\cite{celik2017,wang2019}.

\subsection*{Model standarization for auditing purposes}

Auditing machine learning models is not an easy task. When an auditor wants to audit several models under the same constraints, it is required that all models audited fulfill an equivalent set of requirements. Those requirements may limit the use of certain software libraries, or of certain model architectures. Usually, even within the same company, each model is designed and trained on its own basis. As research in machine learning grows, new models are continuously devised. However, this fast growth in available techniques hinders the possibility of having a deep understanding of those models and makes the assessment of some auditing dimensions a nearly impossible task.

In this scenario, differential replication can be used to establish a small set of canonical models into which all others can be translated. In this sense, deep knowledge of these set of canonical models would be enough to drive auditing tests. For example, let us consider that the canonical model is a deep learning architecture with a certain configuration. Any other model can be translated into this particular architecture using differential replication\footnote{Provided the capacity of the network is large enough to replicate the given decision boundary.}. The auditing process need then only consider how to probe the canonical deep network to report impact assessment.

\section{Conclusions}

In this paper we have described a general framework based on knowledge reuse from generation to generation to extend the useful life of machine learning models by adapting them to their changing environment. To tackle this issue of \textit{environmental adaptation}, we have proposed a mechanism inspired in how biological organisms evolve: \textit{differential replication}. Differential replication allows machine learning models to modify their behavior to meet the new requirements defined by the environment. We envision this replication mechanism as a projection operator able to translate the decision behavior of a machine learning model into a new hypothesis space with different characteristics. This allows traits of a given classifier to be inherited by another, more suitable under the new premises. We have listed different inheritance mechanisms to achieve this goal, depending on specific knowledge availability scenarios. These range from the more permissive \textit{inheritance by sharing the dataset} to the more restrictive \textit{inheritance by copying}, which is the solution requiring less knowledge about the parent model and training data. Finally, we provide examples of how differential replication applies in practice for six different real-life scenarios.

\section*{Acknowledgements}
This work has been partially funded by the Spanish project PID2019-105093GB-I00 (MINECO/FEDER, UE), and by AGAUR of the Generalitat de Catalunya through the Industrial PhD grant 2017-DI-25. 

\bibliographystyle{IEEEtran}
\bibliography{bibliography.bbl}

\end{document}